\documentclass[letterpaper, 10 pt, conference]{ieeeconf}  

\IEEEoverridecommandlockouts                              

\title{\LARGE \bf
ASTER: Attitude-aware Suspended-payload Quadrotor Traversal \\ via Efficient Reinforcement Learning
}
\usepackage{graphicx}
\usepackage{subfigure}
\usepackage{amsmath}
\usepackage{mathtools}
\usepackage{amsfonts,amssymb}
\usepackage{lipsum}
\usepackage{hyperref}
\hypersetup{hidelinks,
	colorlinks=true,
	allcolors=blue,
	pdfstartview=Fit,
	breaklinks=true}
\usepackage{algorithm}
\usepackage{algorithmicx}
\usepackage{algpseudocode}
\usepackage{makecell}
\usepackage{multirow}
\usepackage{cite}
\usepackage{color}
\DeclareMathOperator*{\argmax}{arg\,max}
\usepackage{bm}
\usepackage{tabulary}
\usepackage{threeparttable}
\usepackage{booktabs}
\usepackage{siunitx}
\usepackage{diagbox}
\usepackage{float}

\usepackage[all]{hypcap}

\author{Dongcheng Cao$^{1}$, Jin Zhou$^{1}$, and Shuo Li$^{1}$
\thanks{$^{1}$Authors are with the College of Control Science and Engineering, Zhejiang University, Hangzhou 310027, China.
        {\tt\small cdc11@zju.edu.cn shuo.li@zju.edu.cn}
        }
}

\begin{document}

\maketitle
\thispagestyle{empty}
\pagestyle{empty}


\begin{abstract}
Agile maneuvering of the quadrotor cable-suspended system is significantly hindered by its non-smooth hybrid dynamics. While model-free Reinforcement Learning (RL) circumvents explicit differentiation of complex models, achieving attitude-constrained or inverted flight remains an open challenge due to the extreme reward sparsity under strict orientation requirements. This paper presents \textit{ASTER}, a robust RL framework that achieves, to our knowledge, the first successful autonomous inverted flight for the cable-suspended system. We propose hybrid-dynamics-informed state seeding (HDSS), an initialization strategy that back-propagates target configurations through physics-consistent kinematic inversions across both taut and slack cable phases. HDSS enables the policy to discover aggressive maneuvers that are unreachable via standard exploration. Extensive simulations and real-world experiments demonstrate remarkable agility, precise attitude alignment, and robust zero-shot sim-to-real transfer across complex trajectories.
 [\href{https://www.youtube.com/playlist?list=PLbEQeDMEVpqzsP6haWt2VlvCpcqTf4Fzu}{\textbf{\emph{video}}}\footnote[2]{https://www.youtube.com/playlist?list=PLbEQeDMEVpqzsP6haWt2Vl\-vCpcqTf4Fzu}]
\end{abstract}

\hypersetup{hidelinks,
	colorlinks=true,
	allcolors=black,
	pdfstartview=Fit,
	breaklinks=true}

\section{INTRODUCTION}

The field of aerial robotics has undergone rapid growth in recent years, leading to the development of various platforms with heterogeneous structures \cite{cao2025proximal}, \cite{peng2025dexterous}. Within this domain, the cable-suspended system has gained significant attention. As a structurally elegant yet dynamically complex platform, this system represents a focal point of research with immense potential for various application scenarios \cite{sun2025agile}, \cite{li2024human}.

A primary motivation for research in the cable-suspended system is the continuous exploration of its dynamic potential to facilitate more diverse tasks. Initial studies primarily focus on suppressing cable oscillations to ensure safety, often trading off agility for stability \cite{palunko2012trajectory}, \cite{sreenath2013trajectory}. Subsequent research pushes these boundaries to include navigation through dense obstacle environments and even narrow gates \cite{li2023autotrans}, \cite{wang2024impact}, \cite{sarvaiya2026polyfly}.

\begin{figure}[h!]
    \centering
    \subfigure[Time-lapse sequence of the single-loop maneuver.]
    {\includegraphics[width=0.35\textwidth, trim = 270 115 120 20, clip]{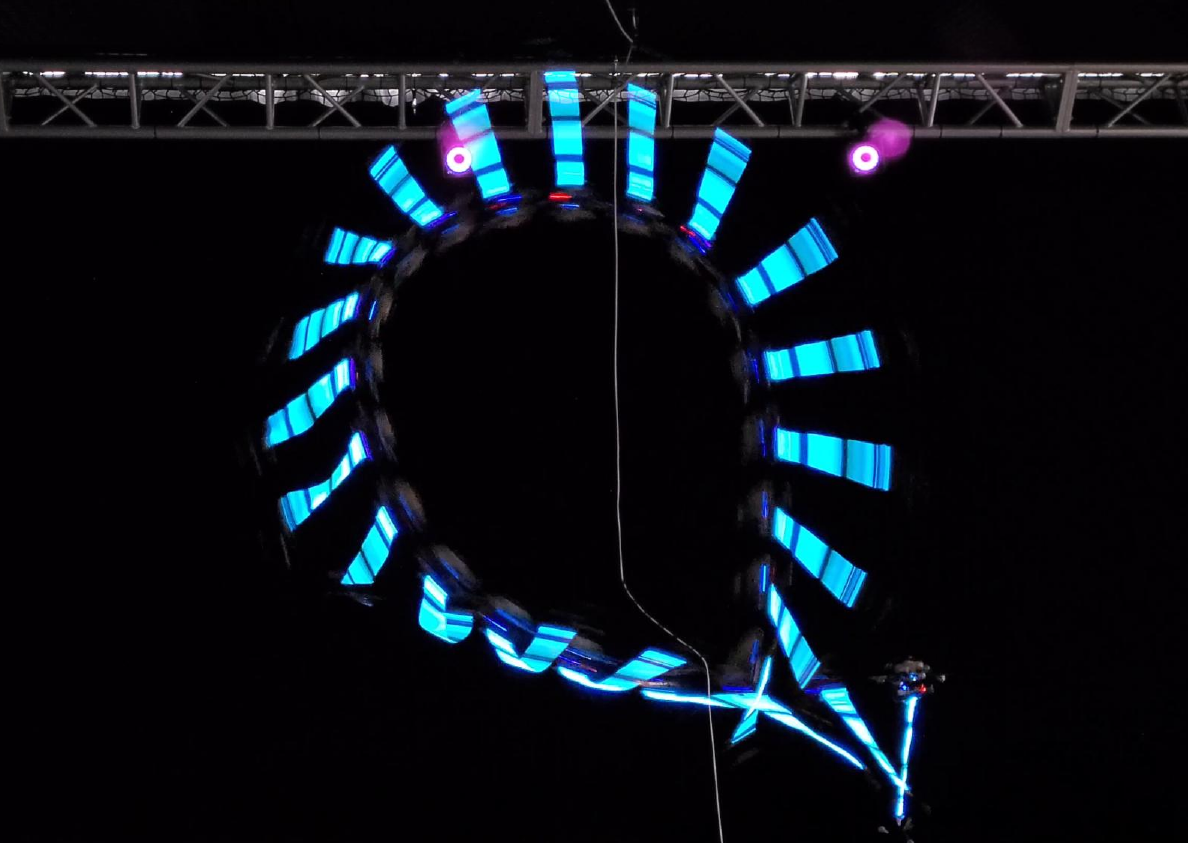}\label{fig: real-world time-lapse-a}}
    \subfigure[Time-lapse sequence of the double-loop maneuver.]{\includegraphics[width=0.35\textwidth, trim = 240 0 0 0, clip]{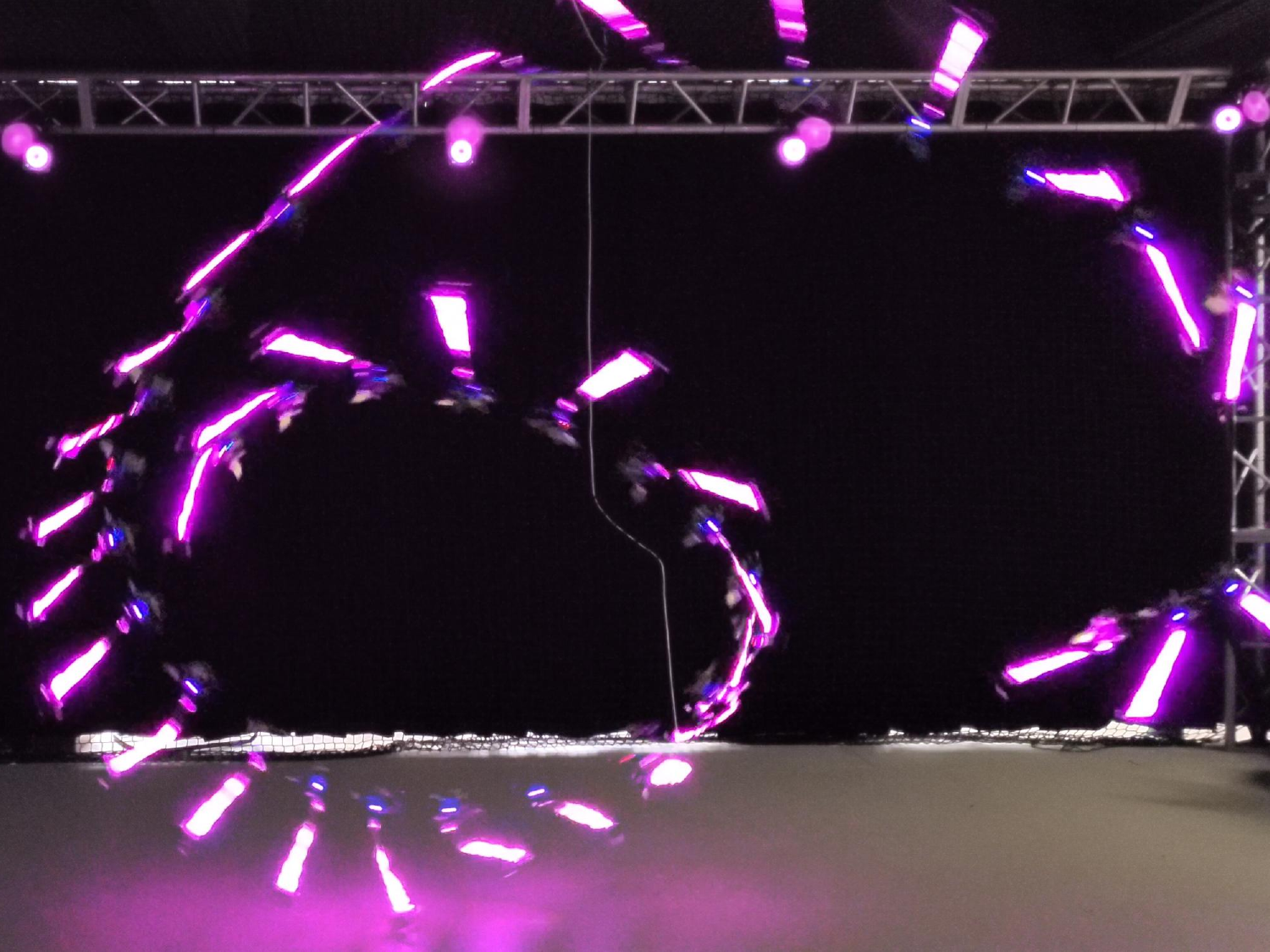}\label{fig: real-world time-lapse-b}}
    \caption{Composite time-lapse illustration of the real-world inverted flight maneuvers, showcasing the continuous motion and attitude transitions.} \label{fig: real-world time-lapse}
    \vspace{-1.3em}
\end{figure}

\begin{figure*}
     \centering
    \includegraphics[width=0.95\textwidth,trim = 0 100 0 0, clip]{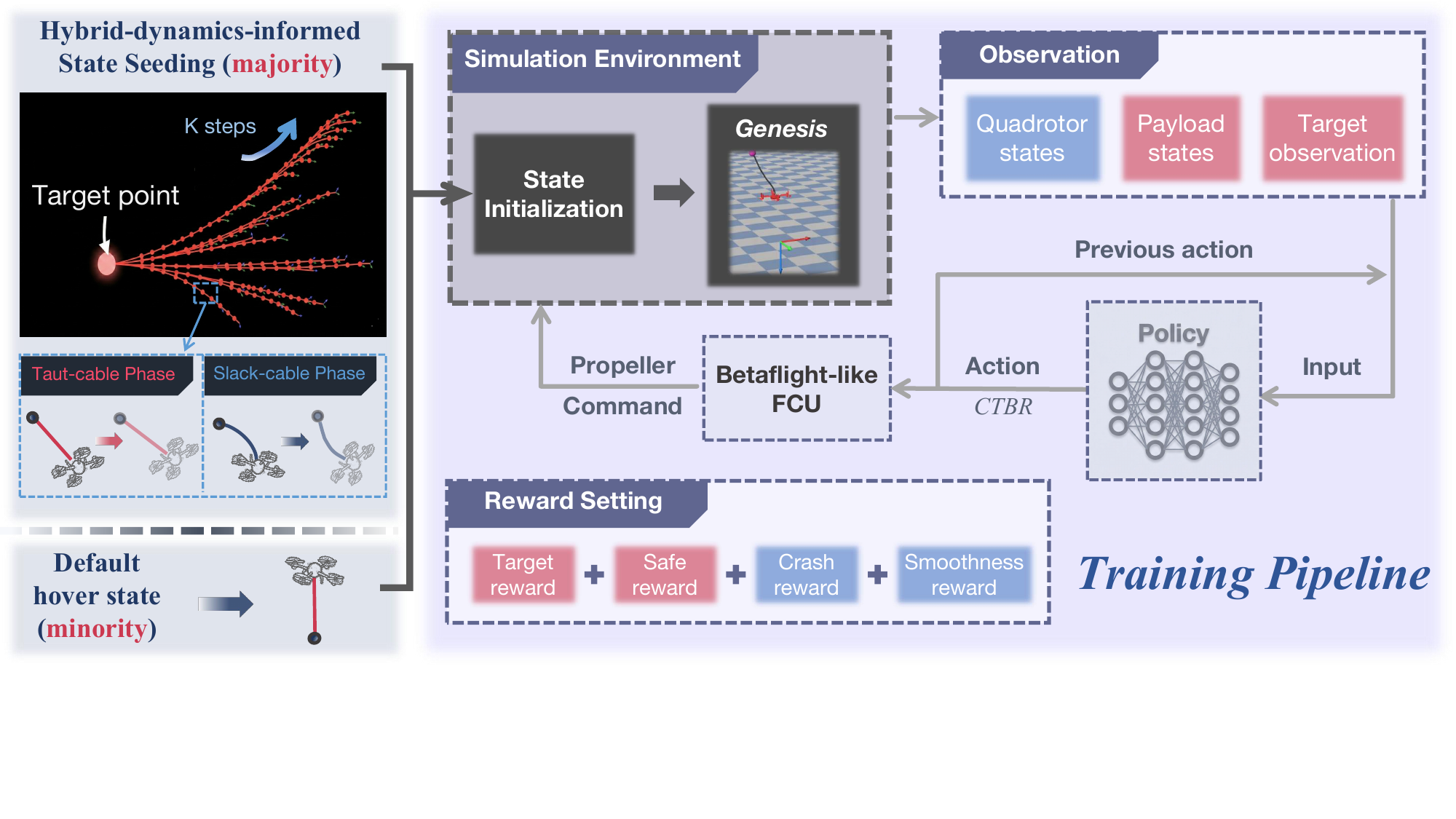}
    \caption{Overview of the training pipeline of the proposed framework. We highlight the design of hybrid-dynamics-informed state seeding (HDSS), a critical strategy for state initialization to overcome exploration bottlenecks in reward-sparse tasks. HDSS back-propagates target configurations through hybrid phases to provide physics-consistent initializations. This strategy balances efficient maneuver discovery with global robustness, enabling the policy to master complex flight dynamics.}
    \label{training pipeline}
    \vspace{-1.2em}
\end{figure*}

In the pursuit of higher maneuverability, the cable-suspended system inevitably encounters hybrid dynamics arising from the transitions between taut and slack cable phases. These inherent non-smooth characteristics significantly impede the computational efficiency of traditional optimization-based methods  \cite{wang2024impact}, \cite{foehn2017fast}. Recently, reinforcement learning (RL) has emerged as a powerful paradigm to enhance quadrotor agility, offering remarkable computational efficiency for real-time onboard deployment and robust sim-to-real transfer \cite{song2023reaching}, \cite{kaufmann2023champion}, \cite{wang2025dashinggoldensnitchmultidrone}. Moreover, model-free RL circumvents the need for explicit differentiation of system dynamics, effectively bypassing the analytical challenges posed by the hybrid dynamics of the cable-suspended system \cite{cao2026flare}. Leveraging these strengths, recent research successfully extend RL to the cable-suspended system, enabling highly agile maneuvers in diverse scenarios \cite{cao2026flare} and complex multi-drone collaborative tasks \cite{zeng2025decentralized}.

Moreover, it is attractive to investigate the attitude-aware tasks for the suspended system, which further unlock the dynamic potential of the system. Parallel to studies on precise quadrotor traversal, reaching targets with prescribed orientations is a common requirement for whole-body control to pass through angled narrow gates \cite{wang2023learning}, \cite{wu2025whole}, or in extreme aerobatic flights that impose even more demanding attitude requirements \cite{wang2025unlocking}, \cite{han2025reactive}, \cite{yin2025taco}.

When employing RL policies to master these maneuvers, such demanding orientation requirements significantly narrow the feasible search space. Notably, recent studies augment RL with the guided reset mechanism \cite{wu2025whole} or curriculum learning strategy\cite{han2025reactive}, \cite{xie2023learning} to facilitate rapid convergence for attitude-aware quadrotor maneuvers, even under the sparse rewards imposed by strict orientation constraints. However, research specifically targeting the cable-suspended system is quite limited. Moreover, the inherent non-smooth hybrid dynamics and underactuated nature of these systems further aggravate the exploration bottleneck. Consequently, training an attitude-aware policy for such systems under sparse rewards and phase transitions remains challenging and unexplored.

Motivated by these challenges, this paper proposes ASTER, a reinforcement learning (RL) framework for the attitude-aware traversal task of the quadrotor cable-suspended system. Specifically, we investigate maneuvers along flight tracks incorporating inverted waypoints, necessitating the satisfaction of exceptionally demanding attitude constraints during flight. By addressing the exploration challenges jointly induced by the demanding attitude requirement and complex system dynamics, our approach facilitates efficient convergence to a high-reward regime. The primary contributions of this work are summarized as follows:
\begin{enumerate}
    \item We present an RL framework that enables the quadrotor cable-suspended system to perform waypoint traversal tasks on random tracks with specific attitude constraints. To the best of our knowledge, this work is the first study to realize inverted flight for this system.

    \item We design the hybrid-dynamics-informed state seeding (HDSS) strategy, which addresses exploration bottlenecks in reward-sparse environments and incorporates the inherent hybrid dynamics of the cable-suspended system into the learning process. This physics-informed approach ensures efficient convergence while further exploiting the dynamic potential of the platform.

    \item We demonstrate the effectiveness of our approach through extensive validations both in simulations and real-world experiments. The proposed method achieves successful sim-to-real transfer, enabling real-time onboard execution of challenging inverted flights.
\end{enumerate}

\section{METHODOLOGY}

\subsection{Suspended-payload system dynamics}
The quadrotor suspended system exhibits hybrid dynamics characterized by two phases depending on whether the cable is taut or slack. 

In the taut phase, the system dynamics is governed by:
\begin{equation}
\begin{aligned}
    &(m_q+m_l)(\mathbf{\ddot{x}}_l-\mathbf{g})=\left(\boldsymbol{\rho}\cdot \mathbf{R}  \mathbf{T}-m_ql(\boldsymbol{\dot{\rho}}\cdot\boldsymbol{\dot{\rho}})\right)\boldsymbol{\rho},\\
    &\mathbf{x}_q=\mathbf{x}_l - l\boldsymbol{\rho},~~\dot{\mathbf{R}}=\mathbf{R} \boldsymbol{\hat\omega},\\
&m_ql\boldsymbol{\ddot \rho} =- m_ql(\boldsymbol{\dot \rho}\cdot \boldsymbol{\dot \rho})\boldsymbol{\rho}+\boldsymbol{\rho}\times (\boldsymbol{\rho} \times \mathbf{R}  \mathbf{T}),\\
&\mathbf{I}_q\boldsymbol{\dot\omega}=\mathbf{M}-\boldsymbol{\omega} \times \mathbf{I}_q \boldsymbol{\omega},
\label{eq:dynamics}
\end{aligned}
\end{equation}
where $\mathbf{T}=[0, 0, T]^\top$ represents the thrust vector, $\mathbf{R}$ denotes the rotation matrix, $\mathbf{g}=[0,0,g]^\top$ is the gravity vector, $l$ is the cable length, $\boldsymbol{\rho}$ denotes the cable direction unit vector from quadrotor to payload. The variables \(\mathbf{x}_q\), \(\boldsymbol{\omega}\), $m_q$, $\mathbf{I}_q$, $\mathbf{M}$ represent the position, body angular rates, mass, the inertia matrix and the moment vector of quadrotor, and \(\mathbf{x}_l\), $m_l$ represent the position and mass of payload, respectively.

In the slack phase, the quadrotor resumes nominal quadrotor dynamics, while the payload obeys free-fall dynamics.

Such complex hybrid dynamics make optimization-based methods computationally expensive. Therefore, this work employs model-free RL to achieve real-time navigation.

\subsection{Problem formulation}
In this work, we aim to enable the cable-suspended system to traverse a sequence of waypoints while strictly adhering to prescribed target attitudes. We formulate the attitude-aware motion planning problem of this system as a general infinite-horizon Markov Decision Process (MDP), defined as \(\mathcal{M}=(\mathcal{S}, \mathcal{A}, \mathcal{P}, \mathcal{R}, \gamma)\). Here, \(\mathcal{S}\) denotes the state space, \(\mathcal{A}\) the action space, \(\mathcal{P}\) the transition probability, \(\mathcal{R}\) the reward function, and \(\gamma\) the discount factor. Within this MDP framework, we aim to optimize the parameters \(\psi\) of a policy network \(\pi_{\psi}\) to maximize the expected discounted return:

\begin{equation}
\pi_{\psi}^{*} = \argmax_{\pi_{\psi}} ~\mathbb{E}\left[ \sum_{t=0}^{\infty} \gamma^t r(t) \right],
\end{equation}
where \(\gamma \in [0,1)\) is the discount factor, and \(r(t)\) represents the immediate reward at time step \(t\). Following this, we define the observations, actions, and reward functions, as the general framework illustrated in Fig. \ref{training pipeline}. For notational brevity, all symbols are expressed in the world frame $\bm W$ unless otherwise specified by a superscript indicating a different reference frame.

\subsubsection{Observations}
The observation vector $\mathbf{o}$ encapsulates the essential system states and target information required for attitude-aware motion planning, formulated as:
\begin{equation}
    \mathbf{o} = [\mathbf{o}_{quad}, \mathbf{o}_{load}, \mathbf{o}_{target}]^\top,
\end{equation}
where $\mathbf{o}_{quad}$, $\mathbf{o}_{load}$, and $\mathbf{o}_{target}$ represent the quadrotor, payload, and target observations, respectively.

Specifically, the quadrotor component $\mathbf{o}_{quad}$ integrates its current states as well as spatial and attitude guidance:
\begin{equation}
    \mathbf{o}_{quad} = [\Delta \mathbf{x}_{q_1}, \Delta \mathbf{x}_{q_2}, \mathbf{v}_q, \operatorname{vec}(\mathbf{R}_{\bm B}^{\bm T})]^\top.
\end{equation}
Among these, $\Delta \mathbf{x}_{q_1}$ and $\Delta \mathbf{x}_{q_2}$ denote the relative position vectors from the quadrotor to the subsequent two waypoints, providing the policy with the necessary look-ahead information. $\mathbf{v}_q \in \mathbb{R}^3$ specifies the quadrotor linear velocity. To represent the attitude error, $\operatorname{vec}(\mathbf{R}_{\bm B}^{\bm T})$ is the flattened relative rotation matrix defined as $\mathbf{R}_{\bm B}^{\bm T} = (\mathbf{R}_{\bm T}^{\bm W})^\top \mathbf{R}_{\bm B}^{\bm W} \in SO(3)$, where $\mathbf{R}_{\bm{T}}^{\bm{W}}$ and $\mathbf{R}_{\bm{B}}^{\bm{W}}$ denote the orientations of the target waypoint frame $\bm{T}$ and the quadrotor body frame $\bm{B}$ relative to the world frame $\bm{W}$, respectively. This formulation projects the quadrotor's body orientation into the target waypoint frame, enabling the policy to perceive the relative orientation discrepancy. 

\begin{figure}[h]
    \centering
    \includegraphics[width=0.36\textwidth, trim={0 320 620 0},clip]{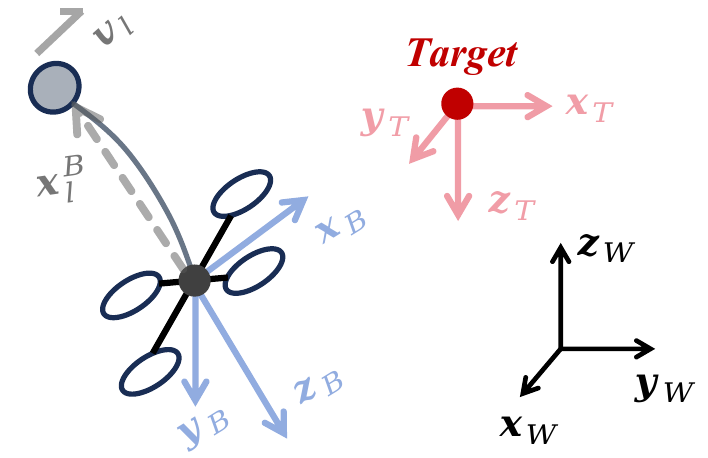}
    \caption{Schematic of the reference frames and payload state: including the world frame $\bm{W}$, the quadrotor body frame $\bm{B}$, and the target frame $\bm{T}$ (illustrated for a waypoint with a downward-pointing $Z$-axis). The vector $\mathbf{x}_l^{\bm{B}}$ denotes the payload position described in the quadrotor body frame.}
    
    \label{fig:coordinate}
\end{figure}

The payload component $\mathbf{o}_{load} = [\mathbf{x}_{l}^{\bm{B}}, \mathbf{v}_l]^\top$ is designed to capture the payload states. It comprises the payload's position vector $\mathbf{x}_{l}^{\bm{B}}$ in body frame $\bm{B}$, as illustrated in Fig. \ref{fig:coordinate}, and its linear velocity $\mathbf{v}_l$ in the world frame. This formulation enables the policy to account for the coupling effects between the quadrotor and the cable-suspended payload.

The target component $\mathbf{o}_{target} = \operatorname{vec}(\mathbf{R}_{\bm T}^{\bm W})$ provides global context via the reference rotation matrix $\mathbf{R}_{\bm T}^{\bm W} \in SO(3)$, which specifies the target's orientation relative to the world frame. The combination of relative and global attitude representations enables the policy to effectively approach the desired maneuver.

\subsubsection{Action Space}
The policy network $\pi_\psi$ outputs a normalized action vector $\mathbf{a} = [\tilde{T}_{\text{cmd}}, \tilde{\boldsymbol{\omega}}_{\text{cmd}}]^\top \in [-1, 1]^4$, representing the collective thrust and body rate setpoints. These normalized actions are then mapped to physical commands through the following linear transformations:
\begin{equation}
    T_{\text{cmd}} = \frac{\tilde{T}_{\text{cmd}} + 1}{2} T_{\text{max}}, \quad
    \boldsymbol{\omega}_{\text{cmd}} = \tilde{\boldsymbol{\omega}}_{\text{cmd}} \odot \boldsymbol{\omega}_{\text{max}},
\end{equation}
where $T_{\text{max}}$ denotes the maximum thrust, $\boldsymbol{\omega}_{\text{max}} \in \mathbb{R}^3$ represents the predefined limits for body rates, and $\odot$ denotes the element-wise product.

\subsubsection{Reward Function}
The reward function is defined as follows:
\begin{equation}
    {r} = {r}_\textit{target}+{r}_\textit{safe}+{r}_\textit{crash}+ {r}_\textit{smooth},
\end{equation}
where we define the sub-rewards as follows.

The target traversal reward $r_{\textit{target}}$ evaluates the precision and success of the traversal under concurrent spatial and attitude constraints. Unlike continuous shaping rewards, this term is granted only when the system satisfies specific spatial and orientation criteria at the moment of traversal. The reward is formulated as a piecewise function:

\begin{equation}
    r_{\textit{target}} = 
    \begin{cases} 
        \lambda_1 e^{-\sigma_p \|\Delta \mathbf{x}_{q_1}\|} + \lambda_2 e^{-\sigma_\theta \theta_{\text{err}}} + C, & \text{if } \text{traversed}, \\
        0, & \text{otherwise},
    \end{cases}
\end{equation}
where $\|\Delta \mathbf{x}_{q_1}\|$ and $\theta_{\text{err}}$ represent the position error and attitude deviation relative to the target waypoint at the moment of traversing, respectively. Specifically, $\sigma_p$ and $\sigma_\theta$ regulate the decay rates, while $\lambda_1, \lambda_2$, and $C$ serve as scaling weights and a constant completion bonus, respectively, with all values detailed in Table \ref{tab:Simulation_Parameters}. Notably, a successful traversal occurs when the following criteria are simultaneously satisfied:
\begin{itemize}
    \item Plane Crossing: The dot product $\Delta \mathbf{x}_{q_1} \cdot \mathbf{x}_{\bm {T}}$ transitions from a negative to a positive value, where $\mathbf{x}_{\bm {T}}$ denotes the X-axis vector in the target frame of the current waypoint. This indicates that the quadrotor has crossed the $Y$-$O$-$Z$ plane of the target frame $\bm {T}$.
    \item Spatial Proximity: The Euclidean distance between the quadrotor and the target is within a predefined threshold, specifically $\| \Delta \mathbf{x}_{q_1} \| < L$, ensuring the quadrotor passes sufficiently close to the waypoint.
    \item Attitude Alignment: The attitude deviation of quadrotor remains within a predefined tolerance, $\theta_{\text{err}} < \epsilon_\theta$, validating the precision of the attitude-aware maneuver.
\end{itemize}

The safety reward $r_{\textit{safe}}$ is designed to mitigate the risk of catastrophic failure during agile flight, specifically addressing the danger of cable-propeller entanglement. Let $z_{l}^{\bm{B}}$ denote the third component of the relative position vector $\mathbf{x}_{l}^{\bm{B}}$ (as defined in the payload observation $\mathbf{o}_{load}$), which represents the payload's vertical displacement along the $Z$-axis of the body frame $\bm{B}$. A penalty is imposed if the payload enters the region above the rotor plane ($z_{l}^{\bm{B}} > 0$). The reward is formulated as follows:

\begin{equation}
    r_{\textit{safe}} = 
    \begin{cases} 
        -r_{exceed}, & z_{l}^{\bm{B}} > 0, \\
        0, & \text{otherwise}.
    \end{cases}
\end{equation}

The crash reward ${r}_{\textit{crash}}$ combines a boundary violation penalty with an immediate termination strategy. Specifically, the crash reward activates exclusively when the system violates its spatially valid workspace $\mathcal{W}$, defined as:
\begin{equation}
    \begin{aligned}
        {r}_{\textit{crash}}=
        \begin{cases}
            -r_{bound}, & \mathbf{x}_q ~\text{or}~ \mathbf{x}_l \notin \mathcal{W}, \\
            0, & \text{otherwise}.
        \end{cases}
    \end{aligned}
\end{equation}

The smoothness reward ${r}_{\textit{smooth}}$ penalizes dynamically infeasible commands through limiting the abrupt variation of consecutive actions $\mathbf{a}$, defined as:
\begin{equation}
    {r}_{\textit{smooth}} =  -\lambda_3 {\left \| \mathbf{a}_{t-1} - \mathbf{a}_{t} \right \|}.
\end{equation}

\subsection{Hybrid-Dynamics-Informed State Seeding}\label{section:hdss}
The simultaneous constraints in the aforementioned $r_{\textit{target}}$ result in extreme reward sparsity, making standard RL prone to non-convergence. To bridge this exploration gap, we propose hybrid-dynamics-informed state seeding (HDSS). Instead of default random resets, HDSS initializes the system in states back-propagated from the given waypoint with its target attitude, effectively accelerating convergence, as illustrated in Fig. \ref{training pipeline}.

The state initialization for each episode is generated by back-propagating the system state $K$ steps prior to reaching a predefined goal state, utilizing a discrete time step of $\Delta t = 0.01\,\text{s}$. Specifically, to start the back-propagation, the quadrotor's goal position is set to the position of the current waypoint. Accordingly, the payload's goal position is randomly sampled from a spherical surface of radius $l$ centered at the quadrotor's goal position, where $l$ denotes the cable length. The remaining components of the goal state are generated via sampling within a prescribed range.

Subsequently, the specific state seeding process between consecutive timesteps is detailed, in which the system states are governed by the following kinematic inversion formula:
\begin{equation}
\begin{aligned}
\boldsymbol{\xi}_{l_{t-1}} &= \mathbf{A}(\Delta t) \boldsymbol{\xi}_{l_t} + \mathbf{B}(\Delta t) \mathbf{s}_{l_{t-1}} + \mathbf{C}(\Delta t),\\
\boldsymbol{\xi}_{q_{t-1}} &= \mathbf{D}(\Delta t) \boldsymbol{\xi}_{q_t} + \mathbf{E}(\Delta t) \mathbf{j}_{q_{t-1}} + \mathbf{F}(\Delta t),
\end{aligned}
\end{equation}
where $t$ denotes the discrete time index within the $K$-step horizon. The payload state vector $\boldsymbol{\xi}_l$ is defined as the vector comprised of the position and its higher-order time derivatives $\boldsymbol{\xi}_l = [\mathbf{x}_l, \mathbf{v}_l, \mathbf{a}_l, \mathbf{j}_l]^\top$, and $\mathbf{s}_l \in \mathbb{R}^3$ represents the payload snap vector. The quadrotor state vector $\boldsymbol{\xi}_q$ is defined as $\boldsymbol{\xi}_q = [\mathbf{x}_q, \mathbf{v}_q, \mathbf{a}_q]^\top$, with $\mathbf{j}_{q}\in \mathbb{R}^3$ representing the quadrotor jerk vector. The definitions of the transition matrices $\mathbf{A}$ through $\mathbf{F}$ vary according to the taut or slack phases of the cable, as formulated below.

\subsubsection{Taut-cable phase}
In the taut-cable phase, according to the kinematic coupling between the compnents of $\boldsymbol{\xi}_l$, we define the matrices $\mathbf{A}$, $\mathbf{B}$, and $\mathbf{C}$ as follows:
\begin{equation}
\begin{aligned}
\mathbf{A}(\Delta t) &= \begin{bmatrix}
1 & -\Delta t & \frac{1}{2}\Delta t^2 & -\frac{1}{6}\Delta t^3 \\
0 & 1 & -\Delta t & \frac{1}{2}\Delta t^2 \\
0 & 0 & 1 & -\Delta t \\
0 & 0 & 0 & 1
\end{bmatrix},\\ \mathbf{B}(\Delta t) &= \begin{bmatrix}
-\frac{1}{24}\Delta t^4 & \frac{1}{6}\Delta t^3 & -\frac{1}{2}\Delta t^2 & \Delta t
\end{bmatrix}^\top, \mathbf{C}(\Delta t) = \mathbf{0}.
\end{aligned}
\end{equation}

Building upon this, the payload state seeding is accomplished with a randomly sampled snap value $\mathbf{s}_{l_{t-1}}$. Similarly, the matrices $\mathbf{D}$, $\mathbf{E}$, and $\mathbf{F}$ are formulated as follows:
\begin{equation}
\begin{aligned}
    \mathbf{D}(\Delta t) &= \begin{bmatrix}
        1 & -\Delta t & \frac{1}{2}\Delta t^2 \\
        0 & 1 & -\Delta t \\
        0 & 0 & 0 \\
    \end{bmatrix} , \quad 
    \mathbf{E}(\Delta t)\mathbf{j}_{q_{t-1}} = \mathbf{0},\\
    \mathbf{F}(\Delta t) &= \begin{bmatrix}
        0 & 0 & \mathbf{a}_{l_{t-1}} + \frac{l}{\| \mathbf{a}_{l_{t-1}} - \mathbf{g} \|} \mathbf{s}_{l_{t-1}}
    \end{bmatrix}^\top,
    \end{aligned}
\end{equation}
which characterizes the state seeding process for the quadrotor, where the matrix $\mathbf{F}(\Delta t)$ is derived by differentiating the position constraint (included in Eq. \ref{eq:dynamics})  between the quadrotor and the payload with respect to time.

Consequently, the quadrotor's orientation is derived from its linear acceleration and a yaw angle aligned with the target frame $\bm{T}$, thereby yielding the full back-propagated states for the state seeding process.

\subsubsection{Slack-cable phase}
In the slack phase, the kinematics of the quadrotor and the payload are decoupled. Specifically, the payload undergoes free-fall motion, and accordingly, the transition matrices $\mathbf{A}$, $\mathbf{B}$, and $\mathbf{C}$ are defined as follows:

\begin{equation}
\begin{aligned}
    \mathbf{A}(\Delta t) &= \begin{bmatrix} 1 & -\Delta t & \frac{1}{2}\Delta t^2 & 0\\ 0 & 1 &  -\Delta t & 0 \\ 0 & 0 & 0 & 0 \\ 0 & 0 & 0 & 0 \end{bmatrix},\quad \mathbf{B}(\Delta t) \mathbf{s}_{l_{t-1}} = \mathbf{0},\\  \mathbf{C}(\Delta t) &= \begin{bmatrix} 0 &  0 & g & 0 \end{bmatrix} ^ \top.
\end{aligned}
\end{equation}

Furthermore, as the quadrotor is no longer subjected to cable tension, we accordingly define the matrices $\mathbf{D}$, $\mathbf{E}$, and $\mathbf{F}$ as follows to complete the state seeding for the quadrotor:
\begin{equation}
\begin{aligned}
    \mathbf{D}(\Delta t) &= \begin{bmatrix} 1 & -\Delta t & \frac{1}{2}\Delta t^2 \\ 0 & 1 & -\Delta t \\ 0 & 0 & 1 \end{bmatrix}, \\ 
    \mathbf{E}(\Delta t) &= \begin{bmatrix} -\frac{1}{6}\Delta t^3 & \frac{1}{2}\Delta t^2 & -\Delta t \end{bmatrix}^\top, \mathbf{F}(\Delta t) = \mathbf{0}.
\end{aligned}
\end{equation}

Notably, to accomplish the slack-cable phase state seeding, we sample the quadrotor jerk $\mathbf{j}_{q_{t-1}}$ instead of the payload snap $\mathbf{s}_{l_{t-1}}$, as the payload experiences constant acceleration in this phase.

\subsubsection{Phase transition}
The state seeding for each episode starts within the taut-cable phase. At each timestep, we monitor for slack transition by examining the term $\mathbf{a}_{l_{t-1}} - \mathbf{g}$. If this value equals zero, implying zero cable tension, the system transitions to the slack-cable phase. 

Correspondingly, during the slack-cable phase, the system monitors for a return to the taut-cable phase by checking if the distance between the quadrotor and payload exceeds the cable length. Once this geometric constraint is violated, the system transitions to the taut-cable phase.

\subsection{Composite State Initialization}
Relying exclusively on initialization with HDSS may hinder the policy's ability to complete flights starting from a default hover state. To achieve efficient exploration with global robustness, the initial reset state is sampled from a non-uniform mixture of the following two functional sets:
\begin{itemize}
    \item \textbf{HDSS Seeds (90\%)}: The majority of resets are initialized from the physics-consistent states generated in Sec.~\ref{section:hdss}. This targeted seeding initializes the system in states with a higher probability of reward acquisition, efficiently facilitating the exploration.
    \item \textbf{Workspace Samples (10\%)}: To maintain global robustness, a minority of episodes are initialized at positions sampled randomly across the workspace $\mathcal{W}$ while the system is set to a default hover state. In these resets, both the quadrotor and payload are initialized with zero linear and angular velocities in an upright orientation. 
\end{itemize}

\subsection{Policy training}
\subsubsection{Algorithm Paradigm and Policy Architecture}
In this paper, we employ Proximal Policy Optimization (PPO) \cite{PPO} to learn the control policy $\pi_\psi$. The policy is parameterized by a multi-layer perceptron (MLP) consisting of two hidden layers with 128 neurons each. The hidden features are mapped to a normalized action vector $\mathbf{a} \in [-1, 1]^4$ via a Tanh activation layer. Moreover, the input observation vector $\mathbf{o}_t$ is augmented with the previous action $\mathbf{a}_{t-1}$.

Before entering the network, all observations are scaled by specific constants, which ensures balanced feature weighting and numerical stability across different units. Specifically, relative positions ($\Delta\mathbf{x}_{q_{1,2}}, \mathbf{x}_{l}^{\bm{B}}$) and linear velocities ($\mathbf{v}_q, \mathbf{v}_l$) are scaled by their respective normalization parameters to align their magnitudes, while the rotation matrices ($\mathbf{R}_{\bm B}^{\bm T}, \mathbf{R}_{\bm T}^{\bm W}$) remain in their original form as their elements are inherently bounded within $[-1, 1]$:
\begin{equation}
\bar{\mathbf{o}} = [ \Delta\mathbf{x}_{q_{1,2}}/\mathbf{k}_q, \mathbf{v}_q/\mathbf{k}_{v}, \mathbf{x}_{l}^{\bm{B}}/\mathbf{k}_l, \mathbf{v}_l/\mathbf{k}_{v}, \operatorname{vec}(\mathbf{R}_{\bm B}^{\bm T}, \mathbf{R}_{\bm T}^{\bm W}) ]^\top,
\end{equation}
where the scaling constants $\mathbf{k}_{\{\cdot\}}$ are detailed in Table~\ref{tab:Simulation_Parameters}.

\begin{table}[htbp]
    \caption{Simulation and training Parameters}
    \begin{center}
        \renewcommand{\arraystretch}{1.2}
        \begin{tabular}{ll|ll}
            \Xhline{1pt}
            \textbf{Parameter} & \textbf{Value} & \textbf{Parameter} & \textbf{Value} \\ \hline
            $T_{\text{max}}$ [m/s$^2$] & $3.5g$ & $\boldsymbol{\omega}_{\text{max}}$ [rad/s] & $[10, 10, 3]$ \\
            $\mathbf{k}_q$ [m] & $[4, 4, 3]$ & $\mathbf{k}_v$ [m/s] & $[5, 5, 5]$ \\ 
            $\mathbf{k}_{l}$ [m] & $[0.5, 0.5, 0.5]$ &  $\sigma_p$  & $3.0$ \\ 
            $\sigma_\theta$ & $2.0$ & $\epsilon_\theta$ [$^\circ$] & 25 \\ 
            $L$ [m] & $0.75$ & $C$ & $5$ \\ 
            $\lambda_1$ & $10$ & $\lambda_2$ & $10$ \\ 
            $\lambda_3$ & $1 \times 10^{\text{-}4}$ & $r_{exceed}$ & $3$ \\
            $r_{bound}$ & $10$ & $K$ & $60$ \\
            $m_q$ [g] & $315$ & $m_l$ [g] & $35$ \\
            $l$ [m] & $0.4$ & $\mathcal{W}$ [m$^3$] & $[-8, 8]^2 \times[0.1, 8]$
            \\ \Xhline{1pt}
        \end{tabular}
    \end{center}
    \label{tab:Simulation_Parameters}
\end{table}

\begin{figure*}[t]
     \centering
    \includegraphics[width=0.98\textwidth,trim = 0 220 0 0, clip]{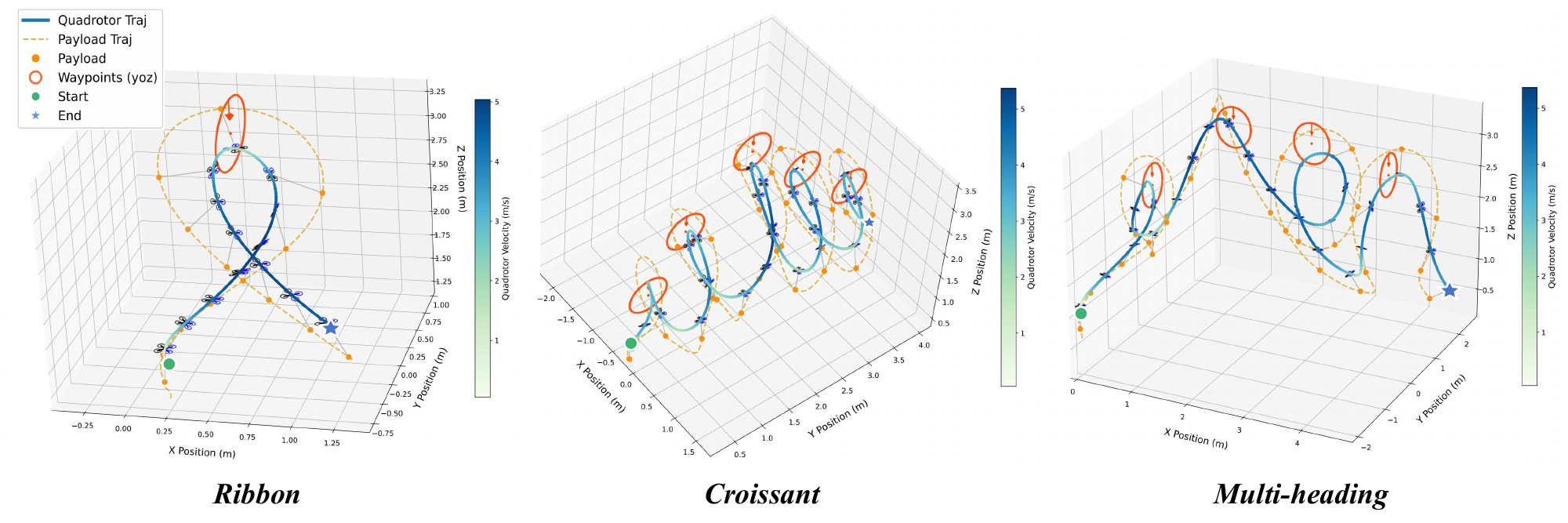}
    \caption{Agile trajectory visualization on 3 representative tracks. The policy is evaluated on (a) \textit{Ribbon}, (b) \textit{Croissant}, and (c) \textit{Multi-heading} tracks. The quadrotor's velocity is visualized via heatmaps, with payload trajectories and system keyframes overlaid. The high-speed, attitude-constrained navigation across all tracks validates the robustness and agility of the learned policy. For visual clarity, only inverted waypoints are denoted by circular markers with embedded arrows indicating the target z-axis direction, while standard upright waypoints are hidden.}
    \label{fig: trajectory in simulation}
    \vspace{-0.3em}
\end{figure*}

\subsubsection{Training Tracks}
The training tracks are characterized by a sequence of target waypoints in $SE(3)$, defined by their spatial positions and target attitudes. These setpoints comprise two distinct types: nominal upright waypoints and inverted waypoints featuring a vertically downward $Z$-axis.

Furthermore, to ensure broad coverage of the state space, target waypoints are dynamically resampled within a relative bounding volume of $[-2, 2] \times [-2, 2] \times [0.5, 3.0]$ m. This resampling is triggered by either a successful traversal or episode termination, exposing the policy to an infinite variety of targets rather than a fixed track.

\subsubsection{Training environment}
Training and evaluation are conducted in Genesis \cite{Genesis}, a GPU-accelerated engine leveraging massive parallelization across thousands of instances. Within this simulator, the cable is represented as a serial chain of small segments, similar to \cite{wang2024impact, cao2026flare, zeng2025decentralized}. By executing all simulations at a timestep of 0.01~s, we achieve a total training volume of $4.1 \times 10^8$ steps. The entire learning process converges in a total training duration of 25 minutes, which is achieved by leveraging 8,192 parallel environments on an NVIDIA RTX 5090 GPU.

\section{SIMULATION RESULTS AND ANALYSIS}
In this section, we evaluate the ASTER framework focusing on three key metrics: maneuver performance, training efficiency, and policy robustness. We first analyze the policy's capability to execute agile maneuvers across diverse demanding tracks. Subsequently, an ablation study is conducted to quantify the necessity of the HDSS strategy in overcoming exploration bottlenecks during training progress. Finally, we assess the system's zero-shot generalization under significant parametric uncertainties, ensuring a solid foundation for sim-to-real deployment.

\subsection{Performance on various tracks}
\begin{figure}[h]
    \centering
    \includegraphics[width=0.36\textwidth, trim={0 0 140 0},clip]{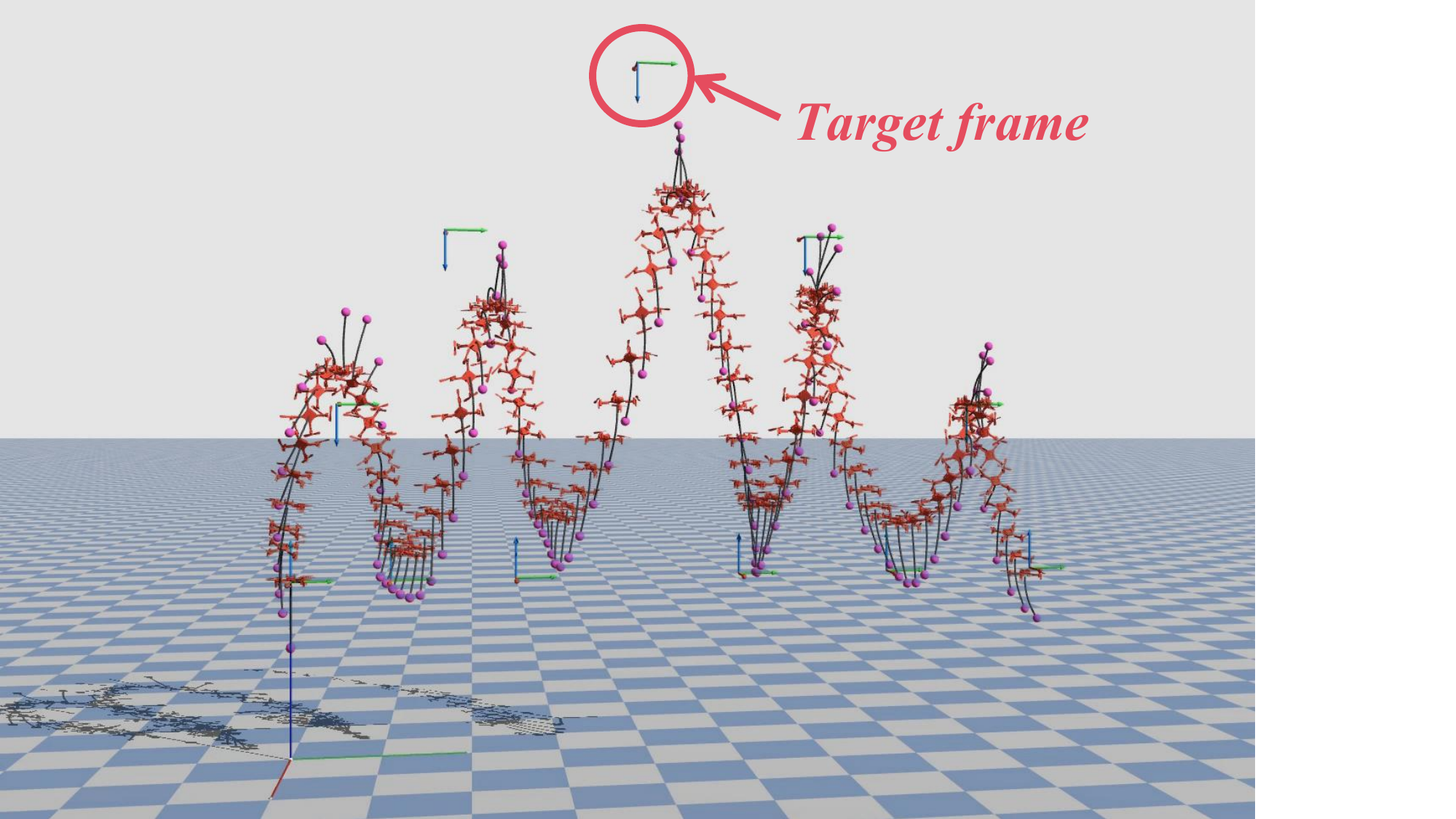}
    \caption{Time-lapse visualization of agile flight on the \textit{Croissant} track in the Genesis simulator. The snapshot sequence demonstrates the policy's capability to execute consecutive inverted maneuvers while maintaining precise attitude alignment and system stability.}
    \label{fig:genesis_traj2}
\end{figure}

We first evaluate the performance of the learned policy across a diverse set of challenging tracks. The cable-suspended system is required to successively traverse a sequence of waypoints defined in $SE(3)$. Notably, these tracks incorporate extremely demanding inverted flight waypoints, where the quadrotor must navigate with its $Z$-axis pointing vertically downward. Such demanding orientation constraints serve to validate the system's capability in handling aggressive, attitude-aware maneuvers.

To provide a detailed analysis within a broad evaluation set, we highlight three representative trajectories from an extensive collection of test tracks. Fig. \ref{fig: trajectory in simulation} illustrates the policy's performance on three custom-designed tracks: \textit{Ribbon}, \textit{Croissant}, and \textit{Multi-heading}, each tailored to evaluate specific aspects of extreme maneuverability. Specifically, \textit{Ribbon} isolates a single inverted segment to examine precision, while \textit{Croissant} consists of consecutive inverted waypoints to test stability under prolonged constraints. Finally, \textit{Multi-heading} features several inverted waypoints with varying yaw angles, necessitating inverted traversals from different directions. The successful execution of these demanding tracks, as evidenced by 3D velocity heatmaps, payload trajectories, and keyframe poses in Fig. \ref{fig: trajectory in simulation}, firmly validates the robustness and agility of the learned policy.

Given the heightened complexity of maintaining stability during prolonged inverted flight, Fig. \ref{fig:genesis_traj2} provides a further time-lapse illustration of the \textit{Croissant} track performed in the Genesis simulator \cite{Genesis}. This visualization captures the system's dynamic capability to execute high-speed maneuvers while maintaining precise attitude alignment.

\begin{table}[h!]
  \centering
  \footnotesize
  \setlength{\tabcolsep}{6pt} 
  \caption{Performance Statistics Across Representative Maneuvering Tracks}
  \label{tab:performance_metrics}
  
  \begin{tabular}{c c S[table-format=2.2] S[table-format=1.2] S[table-format=1.2]}
    \toprule
    \multirow{2}{*}{\textbf{Track}} & \multirow{2}{*}{\textbf{WPs}} & {\textbf{Time}} & {\textbf{Avg Vel.}} & {\textbf{Max Vel.}} \\
    & & {(s)} & {(m/s)} & {(m/s)} \\
    \midrule
    Ribbon         & 3 & 1.61 & 2.77 & 5.03 \\
    Croissant      & 11 & 6.12 & 3.05 & 5.36 \\
    Multi-heading  & 9 & 5.47 & 3.44 & 5.32 \\
    \bottomrule
  \end{tabular}
  
  \vspace{0.5em}
  \raggedright
  \scriptsize 
  Note: WPs denotes the number of waypoints. The velocity-related statistics refer to the quadrotor.
\end{table}

As summarized in Table \ref{tab:performance_metrics} for all three flights, the system successfully completes these attitude-constrained tracks at high speeds, demonstrating that our training framework can extensively exploit the agility potential of the cable-suspended system.

\subsection{Exploration Efficiency Analysis}
To verify the necessity of our HDSS strategy, we conduct an ablation study comparing the full framework against a baseline that utilizes standard initialization which is referred to as ``w/o HDSS". This baseline method resets each episode with the quadrotor and payload in a default hover state at a randomized origin. In contrast, our full framework incorporates HDSS, as detailed in Section \ref{section:hdss}, which seeds each episode with physically consistent, non-zero states back-propagated from the waypoint. All other hyperparameters, network architectures, and reward structures remain identical to ensure a controlled comparison.

\begin{figure}[h!]
    \centering
    \centering
    \includegraphics[width=0.45\textwidth, trim={0 0 0 0},clip]{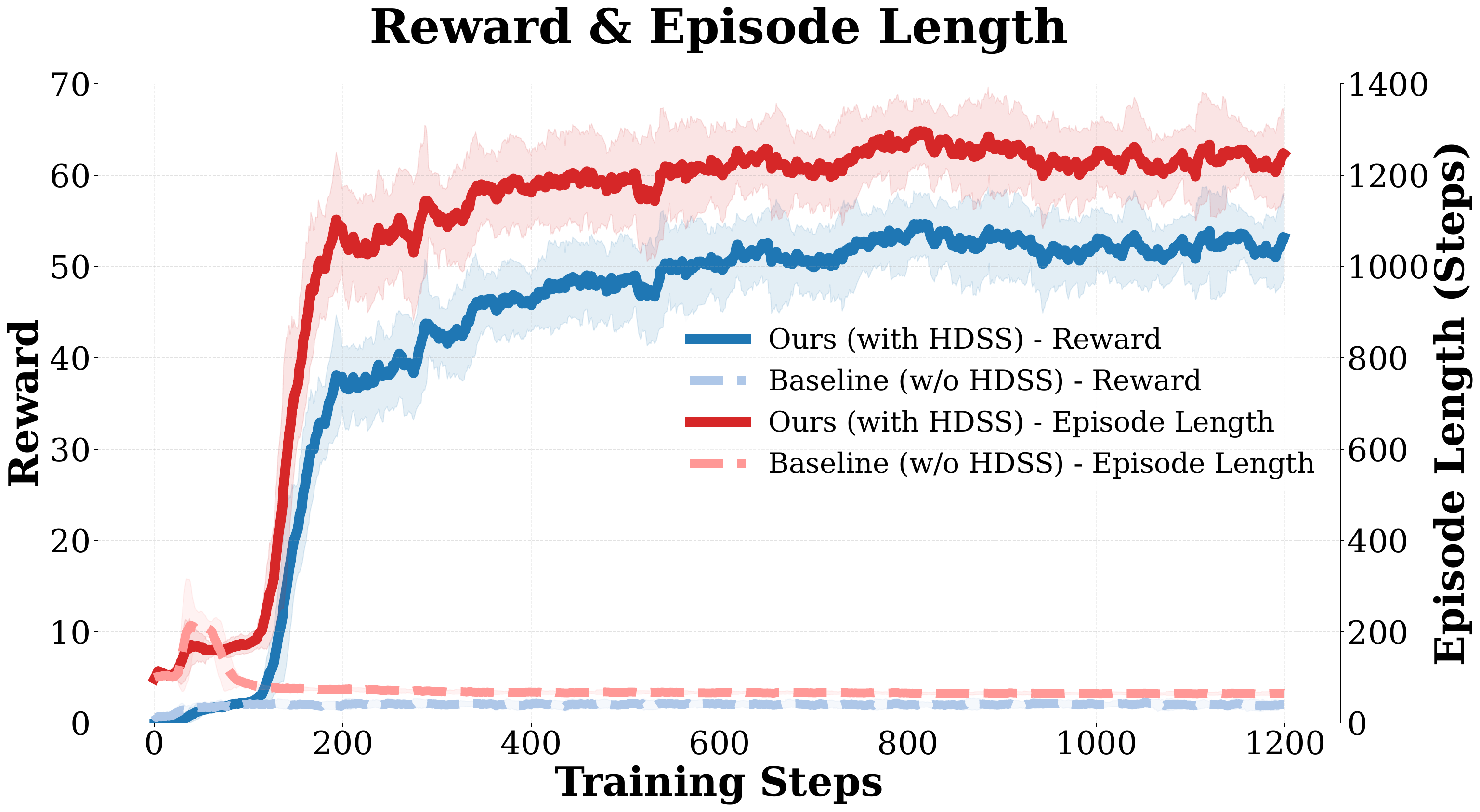}
    \caption{Ablation study of the hybrid-dynamics-informed state seeding (HDSS) strategy. The plots compare the proposed framework against a baseline without HDSS. The baseline remains trapped in a near-zero reward regime, while HDSS enables rapid discovery of high-reward maneuvers.} \label{fig: ablation study}
\end{figure}

The comparative training curves in Fig. \ref{fig: ablation study} illustrate that the baseline policy remains trapped in a near-zero reward regime, highlighting a fundamental exploration bottleneck. This bottleneck is directly mirrored in the episode length evolution, where baseline episodes terminate prematurely due to catastrophic failures or timeouts before reaching the first waypoint.

Conversely, the HDSS-enabled policy demonstrates a rapid and stable increase in both cumulative reward and episode length. By the conclusion of the training phase, the proposed method outperforms the baseline by an order of magnitude across both metrics. This significant performance gap confirms that HDSS is not merely an auxiliary optimization but a critical prerequisite for mastering high-agility inverted maneuvers for the cable-suspended system.

\subsection{Robustness Analysis}
To evaluate the zero-shot generalization of the proposed policy, we conduct robustness tests by varying the payload mass $m_l$ and cable length $l$. While domain randomization (DR) is applied within a $\pm 20\%$ range for both parameters during training, we expand the testing envelope to $\pm 40\%$ to assess the policy's robustness to out-of-distribution states. At each variation point, the evaluation is performed over 200 randomly generated tracks, each containing 10 waypoints. Table \ref{tab:robustness} reports the Success Rates (SR) and Average Completion Times ($T$).

\begin{table}[htbp]
  \centering
  \caption{Robustness Metrics (SR  / $T$ ) under Independent Parameter Variations}
  \label{tab:robustness}
  \renewcommand{\arraystretch}{1.2}
  \setlength{\tabcolsep}{12pt}
  \begin{tabular}{c|cc}
    \hline
    \textbf{Variation} & \textbf{Payload Mass ($m_l$)} & \textbf{Cable Length ($l$)} \\ \hline
    $-40\%$            & 87.5\% / 5.91s                   & 96.5\% / 5.97s                 \\
    $-20\%$            & 93\% / 6.00s                     & 97\% / 6.03s                   \\
    $0\%$ (Nom.)       & 97.5\% / 6.15s                 & 98.5\% / 6.15s                 \\
    $+20\%$            & 91.5\% / 6.28s                   & 82.5\% / 6.22s                 \\
    $+40\%$            & 82\% / 6.47s                     & 56\% / 6.31s                   \\ \hline
  \end{tabular}
  
  \vspace{0.5em}
  \raggedright 
  \scriptsize 
  Note: SR and $T$ denote the success rate and average completion time, respectively.
\end{table}

The quantitative results, summarized in Table \ref{tab:robustness}, validate the robust generalization of the learned policy across a diverse set of random tracks. Within the DR envelope ($\pm 20\%$), the system maintains a high success rate and consistent average completion times $T$, demonstrating that the policy captures the intrinsic varying physics rather than over-fitting to nominal parameters.

Beyond the DR envelope, the policy exhibits different sensitivity to payload mass and cable length. While maintaining a high success rate (above 80\%) under extreme mass variations, the system shows a notable performance disparity regarding the cable length $l$. At the $-40\%$ test point, the success rate still remains high as the reduced swing radius facilitates stable inverted maneuver. Conversely, increasing $l$ to $+40\%$ leads to a drastic drop in success rate. This degradation is primarily due to the increased rotational inertia of the longer cable, which renders the payload significantly harder to maneuver and increases the risk of cable-propeller entanglement during aggressive maneuvers.

\section{EXPERIMENT SETUP AND RESULT}
\subsection{Experimental Setup}
\begin{figure}[h!]
    \centering
    \centering
    \includegraphics[width=0.33\textwidth, trim={0 230 360 0},clip]{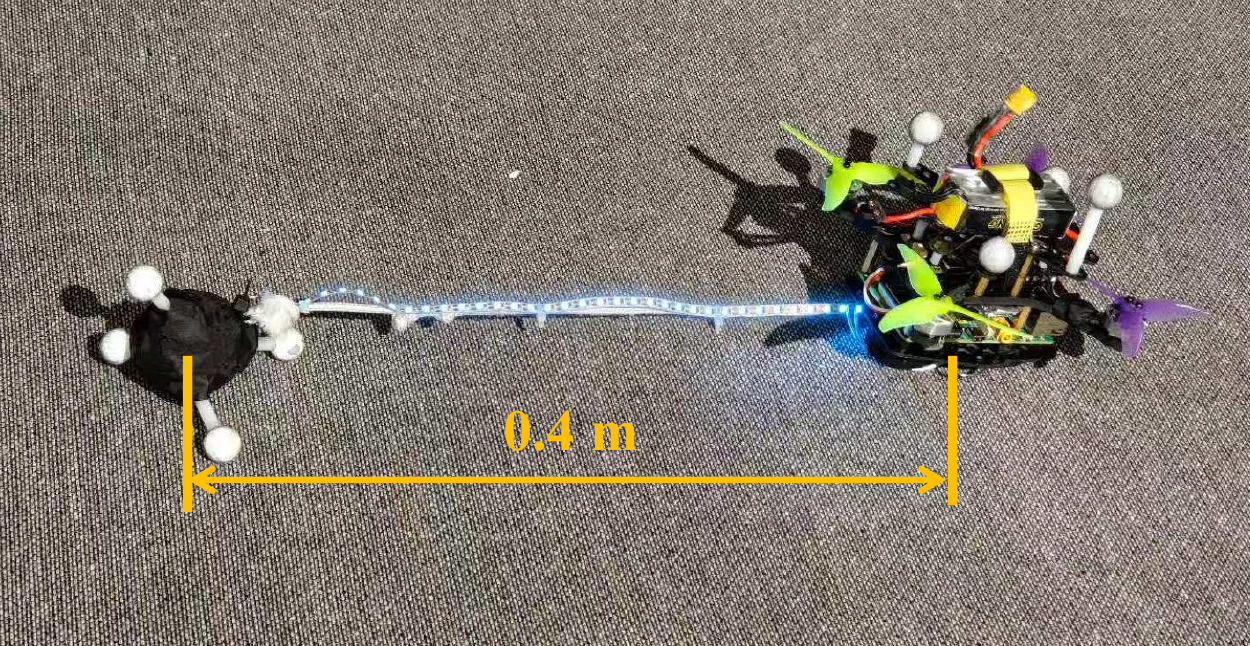}
    \caption{Cable-suspended payload system in real-world experiments.} \label{fig: platform}
\end{figure}

The real-world platform (Fig. \ref{fig: platform}) consists of a \SI{315}{\gram} quadrotor equipped with a \SI{35}{\gram} suspended payload, providing a thrust-to-weight ratio of 3.5. The system employs a hierarchical control architecture: an onboard \textit{Cool Pi} computer executes the RL policy at \SI{100}{\hertz} via \textit{LibTorch}, while a Betaflight-based controller handles low-level rate tracking. A motion capture system provides global states within a \SI{5.5}{\meter} $\times$ \SI{5.5}{\meter} $\times$ \SI{2.5}{\meter} arena. To evaluate the robustness of our method, the policy is transferred zero-shot from simulation to the physical quadrotor without any fine-tuning or domain adaptation.

\subsection{Agile Inverted Maneuvers}
\begin{figure}[h!]
    \centering
    \subfigure[Single-loop 3D reconstructed trajectory.]
    {\includegraphics[width=0.23\textwidth, trim = 0 210 620 0, clip]{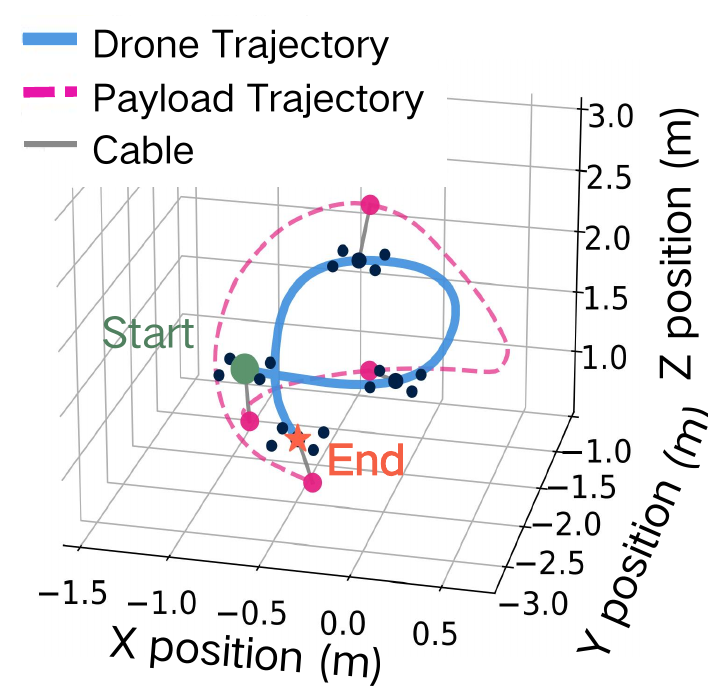}\label{fig: real-world-3D-a}}
    \hspace{0.2cm}
    \subfigure[Double-loop 3D reconstructed trajectory.]{\includegraphics[width=0.23\textwidth, trim = 350 210 270 0, clip]{Figures/loop-real.pdf}\label{fig: real-world-3D-b}}
    \vspace{-0.55em}
    \caption{Reconstructed 3D trajectories of the quadrotor and payload during the agile inverted maneuvers shown in Fig. \ref{fig: real-world time-lapse}} \label{fig: real-world-3D}
\end{figure}

To evaluate the policy's performance in extreme flight regimes, we conduct two representative challenging experiments: a single-loop and a consecutive double-loop inverted flight. The visual results of these maneuvers are captured in Fig. \ref{fig: real-world time-lapse}, while the corresponding quantitative spatial data is illustrated in Fig. \ref{fig: real-world-3D}.

As shown in the single-loop flight (Fig. \ref{fig: real-world time-lapse-a}), the quadrotor successfully executes a vertical loop, reaching a fully inverted pose at the top while preventing the payload from falling into the rotors. The 3D trajectory in Fig. \ref{fig: real-world-3D-a} further confirms the precision of the flight, with the payload (pink dashed line) following a smooth, physics-consistent arc behind the quadrotor.

The double-loop experiment (Fig. \ref{fig: real-world time-lapse-b}) further demonstrates the system's ability to handle rapid, back-to-back transitions between upright and different inverted waypints. The reconstructed 3D data in Fig. \ref{fig: real-world-3D-b} reveals that the system maintains high performance across consecutive inverted waypoints with different positions, validating the robustness and stability of our policy.

\subsection{Sim-to-Real Fidelity}
To evaluate the transferability of the learned policy, we conduct a comparative analysis between the simulation environment and real-world flight on these two identical tracks.

Table \ref{tab:sim_to_real} summarizes the flight statistics, demonstrating that the physical system's performance aligns closely with simulation results. Across both test tracks, maximum velocity discrepancies remain within \SI{6.0}{\percent} with slight discrepancies in completion times. Notably, these discrepancies reduce from the single loop to the double loops, demonstrating the policy's sustained consistency as the task duration scales. Despite these minor variations, the policy successfully stabilizes the hybrid dynamics without real-world fine-tuning, validating the structural robustness of the ASTER framework for zero-shot sim-to-real transfer.
\begin{table}[h]
\centering
\caption{Comparison of Flight Statistics: Simulation vs. Real-world}
\renewcommand{\arraystretch}{1.3}
\label{tab:sim_to_real}
\begin{tabular}{c cccc} 
\toprule
\multirow{2}{*}{\textbf{Track}} & \multicolumn{2}{c}{\textbf{Completion Time (s)}} & \multicolumn{2}{c}{\textbf{Max Quad Vel. (m/s)}} \\
\cmidrule(lr){2-3} \cmidrule(lr){4-5} 
 &\textbf{~~~Sim} &\textbf{~~~Real} &\textbf{~~~Sim} &\textbf{~~~Real} \\
\midrule
Single Loop & ~~~0.98 & ~~~1.34 & ~~~3.83 & ~~~4.01 \\
Double Loops & ~~~2.87 & ~~~3.14 & ~~~7.05 & ~~~6.63\\
\bottomrule
\end{tabular}
\end{table}

\section{CONCLUSIONS}
In this work, we propose ASTER, an RL-based framework for agile, attitude-aware traversal of the quadrotor suspended payload system, specifically focusing on the challenging inverted flight task. The proposed HDSS leverages analytical physics models to guide model-free RL, thereby overcoming persistent exploration bottlenecks. Experimental results demonstrate that the learned policy enables stable inverted maneuvers on random tracks and generalizes to real-world platforms without fine-tuning. Future work will extend the state-seeding strategy to multi-drone suspended systems, facilitating the discovery of policies for complex aerial manipulation or cooperative transport tasks.


\begin{thebibliography}{10}

\bibitem{cao2025proximal}
H.~Cao, J.~Shen, Y.~Zhang, Z.~Fu, C.~Liu, S.~Sun, and S.~Zhao, ``Proximal cooperative aerial manipulation with vertically stacked drones,'' {\em Nature}, vol.~646, no.~8085, pp.~576--583, 2025.

\bibitem{peng2025dexterous}
R.~Peng, Y.~Wang, M.~Lu, and P.~Lu, ``A dexterous and compliant aerial continuum manipulator for cluttered and constrained environments,'' {\em Nature Communications}, vol.~16, no.~1, p.~889, 2025.

\bibitem{sun2025agile}
S.~Sun, X.~Wang, D.~Sanalitro, A.~Franchi, M.~Tognon, and J.~Alonso-Mora, ``Agile and cooperative aerial manipulation of a cable-suspended load,'' {\em Science Robotics}, vol.~10, no.~107, p.~eadu8015, 2025.

\bibitem{li2024human}
G.~Li, X.~Liu, and G.~Loianno, ``Human-aware physical human--robot collaborative transportation and manipulation with multiple aerial robots,'' {\em IEEE Transactions on Robotics}, vol.~41, pp.~762--781, 2024.

\bibitem{palunko2012trajectory}
I.~Palunko, R.~Fierro, and P.~Cruz, ``Trajectory generation for swing-free maneuvers of a quadrotor with suspended payload: A dynamic programming approach,'' in {\em 2012 IEEE international conference on robotics and automation}, pp.~2691--2697, IEEE, 2012.

\bibitem{sreenath2013trajectory}
K.~Sreenath, N.~Michael, and V.~Kumar, ``Trajectory generation and control of a quadrotor with a cable-suspended load-a differentially-flat hybrid system,'' in {\em 2013 IEEE international conference on robotics and automation}, pp.~4888--4895, IEEE, 2013.

\bibitem{li2023autotrans}
H.~Li, H.~Wang, C.~Feng, F.~Gao, B.~Zhou, and S.~Shen, ``Autotrans: A complete planning and control framework for autonomous uav payload transportation,'' {\em IEEE Robotics and Automation Letters}, vol.~8, no.~10, pp.~6859--6866, 2023.

\bibitem{wang2024impact}
H.~Wang, H.~Li, B.~Zhou, F.~Gao, and S.~Shen, ``Impact-aware planning and control for aerial robots with suspended payloads,'' {\em IEEE Transactions on Robotics}, 2024.

\bibitem{sarvaiya2026polyfly}
M.~Sarvaiya, G.~Li, and G.~Loianno, ``Polyfly: Polytopic optimal planning for collision-free cable-suspended aerial payload transportation,'' {\em IEEE Robotics and Automation Letters}, 2026.

\bibitem{foehn2017fast}
P.~Foehn, D.~Falanga, N.~Kuppuswamy, R.~Tedrake, and D.~Scaramuzza, ``Fast trajectory optimization for agile quadrotor maneuvers with a cable-suspended payload,'' 2017.

\bibitem{song2023reaching}
Y.~Song, A.~Romero, M.~M{\"u}ller, V.~Koltun, and D.~Scaramuzza, ``Reaching the limit in autonomous racing: Optimal control versus reinforcement learning,'' {\em Science Robotics}, vol.~8, no.~82, p.~eadg1462, 2023.

\bibitem{kaufmann2023champion}
E.~Kaufmann, L.~Bauersfeld, A.~Loquercio, M.~M{\"u}ller, V.~Koltun, and D.~Scaramuzza, ``Champion-level drone racing using deep reinforcement learning,'' {\em Nature}, vol.~620, no.~7976, pp.~982--987, 2023.

\bibitem{wang2025dashinggoldensnitchmultidrone}
X.~Wang, J.~Zhou, Y.~Feng, J.~Mei, J.~Chen, and S.~Li, ``Dashing for the golden snitch: Multi-drone time-optimal motion planning with multi-agent reinforcement learning,'' 2025.

\bibitem{cao2026flare}
D.~Cao, J.~Zhou, X.~Wang, and S.~Li, ``Flare: Agile flights for quadrotor cable-suspended payload system via reinforcement learning,'' {\em IEEE Robotics and Automation Letters}, 2026.

\bibitem{zeng2025decentralized}
J.~Zeng, A.~M. Gimenez, E.~Vinitsky, J.~Alonso-Mora, and S.~Sun, ``Decentralized aerial manipulation of a cable-suspended load using multi-agent reinforcement learning,'' {\em arXiv preprint arXiv:2508.01522}, 2025.

\bibitem{wang2023learning}
Y.~Wang, B.~Wang, S.~Zhang, H.~W. Sia, and L.~Zhao, ``Learning agile flight maneuvers: Deep se (3) motion planning and control for quadrotors,'' in {\em 2023 IEEE International Conference on Robotics and Automation (ICRA)}, pp.~1680--1686, IEEE, 2023.

\bibitem{wu2025whole}
T.~Wu, Y.~Chen, T.~Chen, G.~Zhao, and F.~Gao, ``Whole-body control through narrow gaps from pixels to action,'' in {\em 2025 IEEE International Conference on Robotics and Automation (ICRA)}, pp.~11317--11324, IEEE, 2025.

\bibitem{wang2025unlocking}
M.~Wang, Q.~Wang, Z.~Wang, Y.~Gao, J.~Wang, C.~Cui, Y.~Li, Z.~Ding, K.~Wang, C.~Xu, {\em et~al.}, ``Unlocking aerobatic potential of quadcopters: Autonomous freestyle flight generation and execution,'' {\em Science Robotics}, vol.~10, no.~101, p.~eadp9905, 2025.

\bibitem{han2025reactive}
Z.~Han, X.~Huang, Z.~Xu, J.~Zhang, Y.~Wu, M.~Wang, T.~Wu, and F.~Gao, ``Reactive aerobatic flight via reinforcement learning,'' {\em IEEE Robotics and Automation Letters}, 2025.

\bibitem{yin2025taco}
Z.~Yin, C.~Zheng, S.~Guo, Z.~Wang, and S.~Zhao, ``Taco: General acrobatic flight control via target-and-command-oriented reinforcement learning,'' in {\em 2025 IEEE/RSJ International Conference on Intelligent Robots and Systems (IROS)}, pp.~9083--9090, IEEE, 2025.

\bibitem{xie2023learning}
Y.~Xie, M.~Lu, R.~Peng, and P.~Lu, ``Learning agile flights through narrow gaps with varying angles using onboard sensing,'' {\em IEEE Robotics and Automation Letters}, vol.~8, no.~9, pp.~5424--5431, 2023.

\bibitem{PPO}
J.~Schulman, F.~Wolski, P.~Dhariwal, A.~Radford, and O.~Klimov, ``Proximal policy optimization algorithms,'' {\em arXiv preprint arXiv:1707.06347}, 2017.

\bibitem{Genesis}
G.~Authors, ``Genesis: A generative and universal physics engine for robotics and beyond,'' December 2024.

\end{thebibliography}
\bibliographystyle{ieeetr}

\end{document}